\documentclass{article}
\usepackage{amsmath,amssymb,amsfonts}
\usepackage{algorithmic}
\usepackage{textcomp}
\usepackage{xcolor}
\usepackage{multirow}
\usepackage{tabularx, booktabs, makecell, multirow, array}
\usepackage{graphicx} 
\usepackage{bm}
\usepackage{svg}
\usepackage{PRIMEarxiv}
\usepackage{subfig}         
\usepackage[utf8]{inputenc} 
\usepackage[T1]{fontenc}    
\usepackage{hyperref}       
\usepackage{url}            
\usepackage{amsfonts}       
\usepackage{nicefrac}       
\usepackage{microtype}      
\usepackage{lipsum}
\usepackage{fancyhdr}       
\graphicspath{{media/}}     
\usepackage[numbers]{natbib}
\title{A Double-Norm Aggregated Tensor Latent Factorization Model for Temporal-Aware Traffic Speed Imputation
}

\author{
  Jiawen Hou \\
  College of Computer and Information Science \\
  Southwest University \\
  Chongqing, China\\
  \texttt{h18761369818@email.swu.edu.cn} \\
   \And
  Hao Wu \\
  College of Computer and Information Science \\
  Southwest University \\
  Chongqing, China\\
  \texttt{haowuf@gmail.com} \\
}

\begin{document}
\maketitle

\begin{abstract}
In intelligent transportation systems (ITS), traffic management departments rely on sensors, cameras, and GPS devices to collect real-time traffic data. Traffic speed data is often incomplete  due to sensor failures, data transmission delays, or occlusions, resulting in missing speed data in certain road segments. Currently, tensor decomposition based methods are extensively utilized, they mostly rely on the $L_2$-norm to construct their learning objectives, which leads to reduced robustness in the algorithms. To address this, we propose Temporal-Aware Traffic Speed Imputation (TATSI), which combines the $L_2$-norm and smooth $L_1$ (${SL}_1$)-norm in its loss function, thereby achieving both high accuracy and robust performance in imputing missing time-varying traffic speed data. TATSI adopts a single latent factor-dependent, nonnegative, and multiplicative update (SLF-NMU) approach, which serves as an efficient solver for performing nonnegative latent factor analysis (LFA) on a tensor. Empirical studies on three real-world time-varying traffic speed datasets demonstrate that, compared with state-of-the-art traffic speed predictors, TATSI more precisely captures temporal patterns, thereby yielding the most accurate imputations for missing traffic speed data.
\end{abstract}


\section{Introduction}
With the continuous advancement of sensor and communication technologies, new traffic sensing devices have been widely deployed in ITS to collect massive spatio-temporal data~\cite{luo2020position}, which would improve both the passenger experience and overall safety and efficiency~\cite{avatefipour2018traffic}. These devices utilize advanced low-latency and highly reliable communication technologies including WiFi 6, low energy Bluetooth, and Ethernet to ensure that geographically~\cite{pascale2012wireless} distributed sensors can transmit and share local sensing data effectively. Additionally, ITS~\cite{zhu2018big} employ various machine learning (ML) methods to analyze the collected data, monitor the overall state of the road traffic network, and adaptively provide multiple functions~\cite{wu2020data}~\cite{yang2023highly}, such as congestion prevention and traffic flow optimization~\cite{tang2024temporal}. However, in practical applications, unpredictable factors such as network transmission interruptions, sensor failures~\cite{luo2021fast}, or storage device malfunctions often result in incomplete traffic data with numerous missing values~\cite{wang2022hybrid}. Consequently, a significant portion of traffic speed data remains unobserved, and the extensive data volume collected over multiple days contributes to its incomplete nature~\cite{wu2022double}. Therefore, accurately and efficiently imputing unobserved traffic speed data remains a vital yet challenging issue~\cite{wu2022prediction}.

Existing methods for missing value imputation typically rely on regression and probabilistic models~\cite{li2020traffic}, including techniques such as singular value decomposition~\cite{liu2020missing}, local least squares regression~\cite{kim2005missing}, probabilistic principal pomponent analysis~\cite{tipping1999probabilistic}, Bayesian PCA~\cite{li2018bpca}, K-nearest neighbors~\cite{batista2003analysis}, and missing forest imputation~\cite{stekhoven2012missforest}. Despite their widespread use, these methods face several challenges~\cite{10734251}. For example, probabilistic models often assume specific data distributions, such as normality, which may not hold in practical scenarios~\cite{wu2020advancing}. Moreover, regression models like local least squares tend to oversimplify complex, high-dimensional data~\cite{hu2020algorithm}, thereby limiting their performance~\cite{luo2021novel}.

Latent factorization of tensors (LFT)-based models have been extensively employed across various tensor analysis tasks. However, most existing LFT methods predominantly rely on the $L_2$-norm~\cite{liu2020convergence}, rendering them susceptible to outliers and significantly compromising their accuracy in data imputation tasks. In contrast, fewer methods utilize the $L_1$-norm~\cite{luo2015nonnegative}. The $L_1$-norm loss, in particular, demonstrates less sensitivity to large errors, thus enhancing the robustness of the models~\cite{leng2019graph}. Moreover, it exhibits smoother behavior for small errors (i.e., errors with absolute values less than 1)~\cite{hu2021effective}, positively contributing to model stability~\cite{yuan2022kalman}. Although LFT models employing $L_2$-norm loss functions can generate stable predictions for missing data in high-dimensional tensor spaces~\cite{koren2021advances}, they generally lack robustness in the presence of outliers~\cite{luo2022neulft}.

From the discussion above, it is evident that neither \(L_1\)-norm nor \(L_2\)-norm loss functions are optimal for modeling LFT models~\cite{wu2021instance}. The \(SL_1\)-norm loss is not only robust to large errors but also exhibits a smooth gradient when the absolute error is less than one~\cite{luo2021alternating}. Therefore, this study adopts a hybrid loss function that integrates both  \(SL_1\)-norm and \(L_2\)-norm losses. In summary, this study makes two primary contributions:
\begin{enumerate}
    \item A TATSI model, which efficiently combines the strong robustness of the \(SL_1\)-norm loss with the stability of the \(L_2\)-norm loss to accurately capture time-varying traffic speed data. Consequently, the model is robust to outliers in high-dimensional and incomplete (HDI) tensors and achieves high prediction accuracy for missing data.
    \item We conduct experiments on three time-varying traffic speed datasets to evaluate the prediction accuracy and of our method.
\end{enumerate}

The rest of the study is organized as follows: Section II describes the preliminaries, Section III proposes a TATSI method, Section IV depicts our experimental results in details, and section V concludes this article.

\section{Preliminaries}
\label{sec:headings}

\subsection{Notations}
\begin{figure}[!htbp]
\centering
\includegraphics[width=0.6\linewidth]{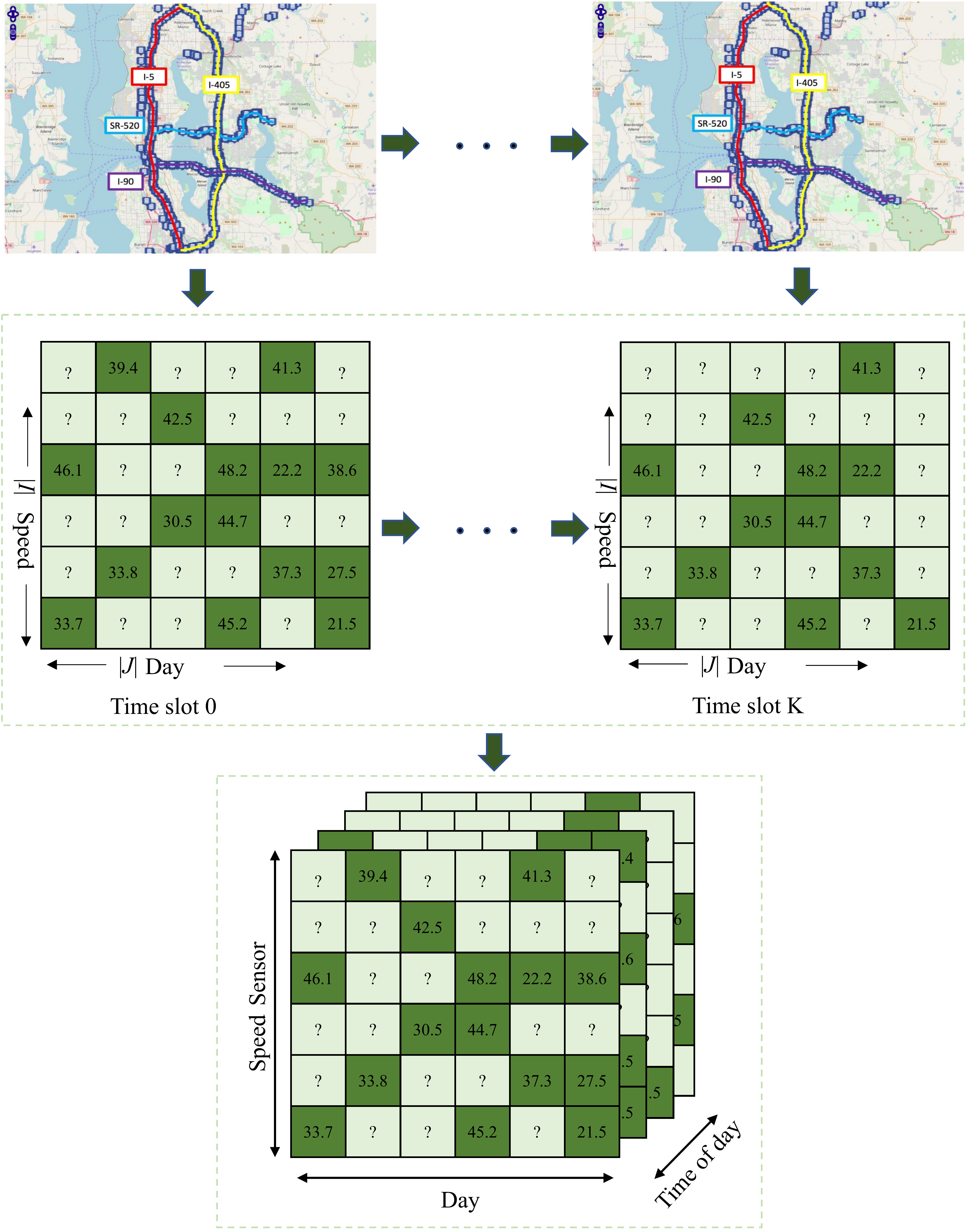}
\caption{An example of the speed sensor-day-time tensor for traffic.}
\label{fig1}
\end{figure}

In this study, we use a third-order traffic speed tensor constructed from observed speed measurements as the primary data source~\cite{li2022momentum}. Since traffic speed data typically consists of non-negative real values, the resulting tensor is also non-negative, as illustrated in Fig.~\ref{fig1}. Notably, the target tensor \(\mathbf{Y}\) usually exhibits high-dimensional incompleteness due to the presence of a large amount of unobserved data~\cite{wu2023dynamic}.

\textit{Definition 1 (Speed sensor-day-time tensor):} Given the constructed tensor \(\mathbf{Y}\), each element \(y_{ijk}\) denotes the observed traffic speed at sensor \(i\in I\), on day \(j\in J\), and at time \(k\in K\), as shown in Fig.~\ref{fig1}. Let \(\Lambda\) denote the set of observed elements and \(\Gamma\) denote the set of unobserved (unknown) elements.

\subsection{Problem statement}
In this study, we apply Canonical/Polyadic Tensor ~\cite{hu2021effective} to represnt the tensor \(\mathbf{Y}\) into its latent factors~\cite{wu2021pid}. Specifically, we express \(\mathbf{Y}\) as a sum of \(R\) rank-one tensors, \(\mathbf{A}_1, \mathbf{A}_2, \ldots, \mathbf{A}_R\), where \(R\) denotes the rank of the approximated tensor \(\hat{\mathbf{Y}}\).

\textit{Definition 2: (Rank-One Tensor):} A tensor $\mathbf{A}_r \in \mathbb{R}^{I \times J \times K}$ is a rank-one tensor which can be out producted by three latent factor (LF) vectors $\mathbf{s}_r$, $\mathbf{d}_r$, and $\mathbf{t}_r$ as $\mathbf{A}_r=\boldsymbol{s}_r\circ\boldsymbol{d}_r\circ\boldsymbol{t}_r$~\cite{luo2024calibrator}.

The LF matrices $\mathrm{S}$, $\mathrm{D}$, and $\mathrm{T}$ consist of \(R\) latent factor vectors with dimensions \(|I|\), \(|J|\), and \(|K|\), respectively, which are combined to form the \(r\)-th rank-one tensor \(\mathbf{A}_r\)~\cite{wu2021pid}. The element-wise expression of \(\mathbf{A}_r\) is obtained by expanding the outer product of these three vectors~\cite{bi2023two}.

\begin{equation}
a_{ijk} = s_{ir}d_{jr}t_{kr}
\end{equation}

To obtain the desired latent factor (LF) matrices $\mathrm{S}$, $\mathrm{D}$, and $\mathrm{T}$, the Euclidean distance is typically employed as a similarity measure to quantify the difference between \(\hat{\mathbf{Y}}\) and  \(\mathbf{Y}\) within the LFT framework~\cite{wu2020advancing}. However, because only a few entries in \(\mathbf{Y}\) are observed, the objective function must be defined on the known dataset \(\Lambda\) to ensure accurate predictions~\cite{yuan2024fuzzy}. Furthermore, non-negativity constraints are imposed on $\mathrm{S}$, $\mathrm{D}$, and $\mathrm{T}$ to preserve their meaningful interpretation~\cite{wu2021discovering}. Therefore, the explicit objective function can be expressed as follows:

\begin{equation}
\begin{aligned}
\begin{gathered}\boldsymbol{\varepsilon}=\sum_{y_{ijk}\in\Lambda}\left\|\mathbf{Y}-\hat{\mathbf{Y}}\right\|^2_{F}
=\sum_{y_{ijk}\in\Lambda}\left(y_{ijk}-\hat{y}_{ijk}\right)^{2}\\
=\sum_{y_{ijk}\in\Lambda}\left(y_{ijk}-\sum_{r=1}^{R}s_{ir}d_{jr}t_{kr}\right)^{2}\\
s.t.\forall i\in I,\forall j\in J,\forall k\in K,r\in\{1,...,R\}:\\s_{ir}\geq0,d_{jr}\geq0,t_{kr}\geq0.\end{gathered}
\end{aligned}
\end{equation}

\textit{Definition 3: (${SL}_1$-norm of a Tensor):} Given a  tensor \(\mathbf{H}\), symbol \(\mathbf{\|H\|}_{{SL_1}}\) denotes the ${SL}_1$-norm of \(\mathbf{H}\) .

\begin{equation}
\label{3}
\mathbf{\|H\|}_{{SL}_1} = 
\begin{cases} 
\displaystyle \sum_{l=1}^L \sum_{m=1}^M \sum_{n=1}^N h_{lmn}^2, & \text{if } |h_{lmn}| \leq 1, \\[8pt]
\displaystyle \sum_{l=1}^L \sum_{m=1}^M \sum_{n=1}^N |h_{lmn}|, & \text{if } |h_{lmn}| > 1 
\end{cases}
\end{equation}

\section{A TATSI method}
\subsection{Objective Function}
To effectively characterize the latent structure of traffic speed tensors and improve prediction accuracy~\cite{yang2024data}, loss functions based on the ${SL}_1$-norm and \(L_2\)-norm each offer distinct advantages for representing HDI data~\cite{goldstein2009split}~\cite{wu2023graph}. Due to the known imbalance in the distribution of \(\mathbf{Y}\) and its sensitivity to the initial assumptions regarding $\mathrm{S}$, $\mathrm{D}$, and $\mathrm{T}$~\cite{liao2025local}, the problem is inherently ill-posed~\cite{zeng2024fast}. Consequently, it is essential to form a linear combination of the ${SL}_1$-norm and the \(L_2\)-norm to balance the model's generality and robustness~\cite{wu2021proportional}. To fully integrate these benefits, this paper proposes an objective function that combines the ${SL}_1$-norm and \(L_2\)-norm, as follows:

\begin{equation}
\label{4}
\begin{aligned} 
\begin{gathered} \varepsilon=\|\Delta\|_{SL_1}^2+\|\Delta\|_{L_2}^2+\lambda\left(\|\mathrm{{S}}\|_{F}^2+\|\mathrm{{D}}\|_{F}^2+\|\mathrm{{T}}\|_{F}^2\right)\\  =\sum_{y_{ijk}\in\Lambda}\left(\left(\Delta_{ijk}\right)_{SL_1}^2+\left(\Delta_{ijk}\right)_{L_2}^2+\lambda
\sum_{r=1}^{R}\left (s_{ir}^2+d_{jr}^2+t_{kr}^2\right)\right)\\ 
s.t.\forall i\in I,\forall j\in J,\forall k\in K,r\in\{1,\ldots,R\}:\\  s_{ir}\geq0,d_{jr}\geq0,t_{kr}\geq0.
\end{gathered}
\end{aligned}
\end{equation}

The operator \(\|\cdot\|_{F}\) computes the Frobenius norm between the observed and imputed values~\cite{zhong2024alternating}, thereby balancing the model’s fit to the traffic speed data~\cite{chen2021efficient}.The norm \(\|\cdot\|_{SL_1}\) represents the ${SL}_1$-norm, which promotes sparsity in the factor matrices through a sparse regularization term~\cite{chen2024sdgnn}, thereby suppressing noise and enhancing the model’s robustness~\cite{luo2021adjusting}.

where \(\Delta_{ijk}\) denotes the prediction error between \(y_{ijk}\) and \(\hat{y}_{ijk}\), defined as:

\begin{equation}
\Delta_{ijk}=y_{ijk}-\hat{y}_{ijk}=y_{ijk}-\sum_{r=1}^{R}s_{ir}d_{jr}t_{kr}.
\end{equation}

Here, \(\mathbf{Y}\) represents the observed traffic speed tensor containing missing values, while \(\hat{\mathbf{Y}}\) denotes the reconstructed tensor obtained via decomposition~\cite{wang2024distributed}~\cite{hu2023fcan}. The factor matrices $\mathrm{{S}}$, $\mathrm{{D}}$, and $\mathrm{{T}}$ correspond, respectively, to speed sensors, days, and time of day~\cite{jiang2024iterative}.

Due to ${SL}_1$-norm behaves differently when \(|\Delta_{ijk}|\le 1\) versus \(|\Delta_{ijk}|>1\), we rewrite the objective function (\ref{4}) in a piecewise form to more accurately characterize the impact of errors on the model~\cite{qin2023asynchronous}.

\begin{equation}
\begin{aligned}
\begin{gathered} 
\epsilon =
\begin{cases}
\displaystyle
\sum_{y_{ijk} \in \Lambda}
\Bigl(
  2(\Delta_{ijk})^2
  + \lambda \sum_{r=1}^{R} \bigl(s_{ir}^2 + d_{jr}^2 + t_{kr}^2\bigr)
\Bigr), &\text{if } |\Delta_{ijk}| \le 1,\\[6pt]
\displaystyle
\sum_{y_{ijk} \in \Lambda}
\Bigl(
  |\Delta_{ijk}| + (\Delta_{ijk})^2
  + \lambda \sum_{r=1}^{R} \bigl(s_{ir}^2 + d_{jr}^2 + t_{kr}^2\bigr)
\Bigr),&\text{if } |\Delta_{ijk}| > 1,
\end{cases}\\[6pt]
\text{s.t.}
\forall i \in I,\; \forall j \in J,\; \forall k \in K,\; \forall r \in \{1, \ldots, R\}:\;\\
s_{ir} \ge 0,\; d_{jr} \ge 0,\; t_{kr} \ge 0.
\end{gathered}
\end{aligned}
\end{equation}

\subsection{Parameter Learning Procedure}
Building upon the objective function discussed above, we introduce a single latent factor-dependent, nonnegative, and multiplicative update (SLF-NMU) strategy~\cite{luo2014efficient}. This method updates parameters via multiplicative adjustments at each iteration, thereby ensuring nonnegativity and accelerating convergence~\cite{fang2024modularity}.

Within the gradient descent framework, the partial derivatives of \(\varepsilon\) with respect to \(s_{ir}\), \(d_{jr}\), and \(t_{kr}\) are computed~\cite{qin2023adaptively}~\cite{bi2023proximal}. The general form of the gradient update can be expressed as follows:

\begin{equation}
\label{Eq2}
(\mathrm{S},\mathrm{D},\mathrm{T}) = \mathop{\arg\min}\limits_{\mathrm{S},\mathrm{D},\mathrm{T}}\varepsilon
\Rightarrow
\left\{
\begin{aligned}
s_{ir} &\leftarrow s_{ir} - \eta_{ir}\frac{\partial\varepsilon}{\partial s_{ir}} \\[6pt]
d_{jr} &\leftarrow d_{jr} - \eta_{jr}\frac{\partial\varepsilon}{\partial d_{jr}} \\[6pt]
t_{kr} &\leftarrow t_{kr} - \eta_{kr}\frac{\partial\varepsilon}{\partial t_{kr}}
\end{aligned}
\right.
\end{equation}

Since the objective function employs a piecewise ${SL}_1$-norm for \(\Delta_{ijk}\), the error is handled differently across various intervals~\cite{chen2024robust}. Consequently, the corresponding gradient must also be computed in a piecewise manner. The partial derivative with respect to \(s_{ir}\) is shown in (\ref{8}).

\begin{equation}
\label{8}
\begin{gathered}
\frac{\partial \epsilon}{\partial s_{ir}} =
\begin{cases}
\displaystyle
\sum_{y_{ijk} \in \Lambda(i)} 
\Bigl(
  d_{jr}t_{kr}
  - 2\bigl(y_{ijk} - \hat{y}_{ijk}\bigr)\,d_{jr}t_{kr}
  + 2\,\lambda\,s_{ir}
\Bigr),&\text{if } \Delta_{ijk} < -1,\\[8pt]
\displaystyle
\sum_{y_{ijk} \in \Lambda(i)} 
\Bigl(
  -4\bigl(y_{ijk} - \hat{y}_{ijk}\bigr)\,d_{jr}t_{kr}
  + 2\,\lambda\,s_{ir}
\Bigr), &\text{if } \lvert \Delta_{ijk} \rvert \le 1,\\[8pt]
\displaystyle
\sum_{y_{ijk} \in \Lambda(i)} 
\Bigl(
  -\,d_{jr}t_{kr}
  - 2\bigl(y_{ijk} - \hat{y}_{ijk}\bigr)\,d_{jr}t_{kr}
  + 2\,\lambda\,s_{ir}
\Bigr), &\text{if } \Delta_{ijk} > 1.
\end{cases}
\end{gathered}
\end{equation}

Here, \(\eta_{ir}\) denotes the learning rate associated with the latent factor in the \(i\)-th dimension~\cite{liu2024hp}. Due to the presence of negative terms in the formulation, latent factors may become negative during training~\cite{zeng2024novel}. To strictly enforce non-negativity, we adopt specific learning rate settings for each latent factor that eliminate negative values~\cite{wang2024dynamically}. For simplicity, we present the derivation for $\mathrm{{S}}$; similar derivations apply to $\mathrm{{D}}$ and $\mathrm{{T}}$~\cite{li2023saliency}. The updates are performed as follows:

\begin{equation}
\label{Eq3}
\left
\{
\begin{aligned}
&\eta_{ir} = \frac{s_{ir}}{\displaystyle\sum_{y_{ijk}\in \Lambda(i)} \left(d_{jr}t_{kr} + 2\hat{y}_{ijk}d_{jr}t_{kr} + 2\lambda s_{ir}\right)}, 
&\text{if }\Delta_{ijk}<-1, \\
&\eta_{ir} = \frac{s_{ir}}{\displaystyle\sum_{y_{ijk}\in \Lambda(i)} \left(4\hat{y}_{ijk}d_{jr}t_{kr} + 2\lambda s_{ir}\right)},
&\text{if } \left|\Delta_{ijk}\right|\leq 1, \\
&\eta_{ir} = \frac{s_{ir}}{\displaystyle\sum_{y_{ijk}\in \Lambda(i)} \left(2\hat{y}_{ijk}d_{jr}t_{kr} + 2\lambda s_{ir}\right)},
&\text{if } \Delta_{ijk}>1.\\
\end{aligned}
\right.
\end{equation}

Substituting (\ref{8}) and (\ref{Eq3}) into (\ref{Eq2}) and simplifying, we obtain the following update rule (\ref{Eq4}):

\begin{equation}
\label{Eq4}
s_{ir} \leftarrow 
\left\{
\begin{aligned}
&s_{ir}\frac{\displaystyle\sum_{y_{ijk}\in\Lambda(i)}2y_{ijk}d_{jr}t_{kr}}{\displaystyle\sum_{y_{ijk}\in\Lambda(i)}\left(d_{jr}t_{kr}+2\hat{y}_{ijk}d_{jr}t_{kr}+2\lambda
s_{ir}\right)}, &\text{if } \Delta_{ijk}<-1, \\[8pt]
&s_{ir}\frac{\displaystyle\sum_{y_{ijk}\in\Lambda(i)}4y_{ijk}d_{jr}t_{kr}}{\displaystyle\sum_{y_{ijk}\in\Lambda(i)}\left(4\hat{y}_{ijk}d_{jr}t_{kr}+2\lambda s_{ir}\right)},&\text{if}\left|\Delta_{ijk}\right|\leq 1, \\[8pt]
&s_{ir}\frac{\displaystyle\sum_{y_{ijk}\in\Lambda(i)}\left(d_{jr}t_{kr}+2y_{ijk}d_{jr}t_{kr}\right)}{\displaystyle\sum_{y_{ijk}\in\Lambda(i)}\left(2\hat{y}_{ijk}d_{jr}t_{kr}+2\lambda s_{ir}\right)},&\text{if} \Delta_{ijk}>1.
\end{aligned}
\right.
\end{equation}

\section{Empirical studies}

\subsection{Experimental Setup}

\textbf{Datasets:}This study employs three publicly available traffic speed datasets. D1 was collected by inductive loop detectors deployed on highways in the Seattle area, USA~\cite{cui2018deep}. A total of 323 loop detectors were selected, recording traffic speeds every 5 minutes, resulting in 288 measurements per sensor per day~\cite{cui2019traffic}. Thus, the traffic data is organized as tensor $\textbf{A}^{(1)} \in \mathbb{R}^{323 \times 288 \times \ 288}$. D2 comprises traffic speed data collected from 214 anonymous road segments (mainly urban expressways and arterial roads) in Guangzhou~\cite{chen2018spatial}, China, over two months (61 days, from August 1 to September 30, 2016), with intervals of 10 minutes. Thus, the traffic data is organized as tensor $\textbf{A}^{(2)} \in \mathbb{R}^{214 \times 61 \times 144}$. D3 consists of traffic speed data collected in urban areas with 18 road segments over 28 days, with a sampling interval of 5 minutes (resulting in 288 time windows per day). Thus, this traffic data is organized as tensor $\textbf{A}^{(3)} \in \mathbb{R}^{18 \times 28 \times 288}$. In order to ensure a robust evaluation of the proposed model and to mitigate overfitting, each dataset is partitioned in two ways for training, validation, and testing: 10\%:20\%:70\% and 20\%:20\%:60\%, respectively, which are shown in TABLE \ref{tab:datasets}.


\begin{table}[!htbp]
\centering
\caption{Datasets}
\label{tab:datasets}
\renewcommand\arraystretch{1.2}
\begin{tabularx}{\linewidth}{>{\centering\arraybackslash}p{1.2cm} 
                                   >{\centering\arraybackslash}p{1.2cm} 
                                   >{\centering\arraybackslash}p{2.8cm} 
                                   >{\centering\arraybackslash}p{2.2cm} 
                                   >{\centering\arraybackslash}p{1.6cm} 
                                   >{\centering\arraybackslash}p{1.6cm} 
                                   >{\centering\arraybackslash}p{1.6cm}}
\toprule
\multicolumn{2}{c}{\textbf{Datasets}} & 
\textbf{|\(\Omega\)|:|\(\Psi\)|:|\(\Phi\)|} & 
\textbf{Total Entries} & 
\makecell{\textbf{Sensor} \\ \textbf{Count}} & 
\makecell{\textbf{Day} \\ \textbf{Count}} & 
\makecell{\textbf{Time} \\ \textbf{Count}} \\
\midrule
\multirow{2}{*}{\textbf{D1}} 
  & D1.1 & {10\%:20\%:70\%} & 92,885 & 323 & 28 & 288 \\
  & D1.2 & {20\%:20\%:60\%} & 92,885 & 323 & 28 & 288 \\
\addlinespace
\multirow{2}{*}{\textbf{D2}} 
  & D2.1 & {10\%:20\%:70\%} & 49,925 & 214 & 61 & 144 \\
  & D2.2 & {20\%:20\%:60\%} & 49,925 & 214 & 61 & 144 \\
\addlinespace
\multirow{2}{*}{\textbf{D3}} 
  & D3.1 & {10\%:20\%:70\%} & 92,885 & 18 & 28 & 288 \\
  & D3.2 & {20\%:20\%:60\%} & 92,885 & 18 & 28 & 288 \\
\bottomrule
\end{tabularx}
\end{table}

\textbf{Evaluation Metrics:} We emphasize the accuracy of traffic speed imputations, as it directly indicates whether the model has captured the essential characteristics of an HDI tensor~\cite{wu2021neural}. Consequently, we employ the mean absolute error (MAE) and root mean square error (RMSE) as evaluation metrics. Let \(\hat{y}_{ijk}\) and \(y_{ijk}\) denote the estimated and actual values, respectively~\cite{luo2021fast}. The two expressions are defined as follows~\cite{luo2023predicting}:

\begin{equation}
\begin{aligned}
\text{RMSE} &= \sqrt{\frac{1}{|\Phi|} \sum_{y_{ijk} \in \Phi} (y_{ijk} - \hat{y}_{ijk})^2} \\
\text{MAE} &= \frac{1}{|\Phi|} \sum_{y_{ijk} \in \Phi} \left| y_{ijk} - \hat{y}_{ijk} \right|
\end{aligned}
\end{equation}

For a tested model, small RMSE and MAE value express high prediction accuracy.

\textbf{Experiment Details:} To ensure a fair evaluation of model performance, all models were implemented on an i7 CPU in a Java (JDK 1.8) environment. Convergence is determined when the loss difference between two consecutive iterations falls below \(10^{-5}\) or when the number of iterations exceeds 1000.

\textbf{Baselines:} To evaluate the performance of the proposed TATSI, we compare it against the following state-of-the-art models: \(M2\) (a quality-of-service prediction model employing the classic Frobenius loss without bias~\cite{luo2019temporal}), \(M3\) (an integrated environment QoS prediction model~\cite{su2021tensor}), and \(M4\) (a QoS prediction model utilizing the Cauchy loss~\cite{ye2021outlier}). To ensure a fair comparison, the latent feature dimension \(R\) for all models is set to 20.

\textbf{Hyperparameter Settings:} The hyperparameter that affects the performance of mine model is the regularization coefficient $\lambda$. We search $\lambda$ from the range [$2^{-20}$,$2^{0}$] with a step size of $2^{-1}$, and finally decide to set $\lambda=9.765625\times 10^{-4}$ on $D1$ and $D2$, and set $\lambda=1.0$ on $D3$. For the hyperparameters of other baselines, they are fine-tuned from the settings in their papers.

\subsection{Experimental Results and Analysis}\label{SCM}
Table~II records the RMSE and MAE of M1--M4 on four cases. Fig.~2 and Fig.~3 show the RMSE and MAE of all models. From the results, we conduct the following analysis:

\begin{table}[!htbp]
\centering
\caption{RMSE and MAE of M1-M4 on datasets}
\label{tab:results}
\begin{tabularx}{\linewidth}{X X X X X X X}
\toprule
\multicolumn{2}{c}{\textbf{Datasets}} & \textbf{M1} & \textbf{M2} & \textbf{M3} & \textbf{M4}  \\
\midrule
\multirow{2}{*}{\textbf{D1.1}} 
    & \textbf{RMSE} & \textbf{5.5488} & 5.6154 & 5.8454 & 5.5947  \\
    & \textbf{MAE}  & \textbf{3.4083} & 3.4829 & 3.6865 & 3.4694  \\
\addlinespace
\multirow{2}{*}{\textbf{D1.2}} 
    & \textbf{RMSE} & \textbf{5.4963} & 5.5039 & 5.5322 & 5.5008  \\
    & \textbf{MAE}  & \textbf{3.3782} & 3.4132 & 3.4624 & 3.3980  \\
\addlinespace
\multirow{2}{*}{\textbf{D2.1}} 
    & \textbf{RMSE} & \textbf{4.2949} & 4.3139 & 4.6317 & 4.3179  \\
    & \textbf{MAE}  & \textbf{2.8414} & 2.8799 & 3.1242 & 2.8765  \\
\addlinespace
\multirow{2}{*}{\textbf{D2.2}} 
    & \textbf{RMSE} & \textbf{4.2476} & 4.3356 & 4.3540 & 4.2378  \\
    & \textbf{MAE}  & \textbf{2.7841} & 2.8052 & 2.9043 & 2.7967  \\
\addlinespace
\multirow{2}{*}{\textbf{D3.1}} 
    & \textbf{RMSE} & \textbf{9.9916} & 10.2205 & 10.8299 & 10.4492  \\
    & \textbf{MAE}  & \textbf{7.0067} & 7.2274 & 7.7656 & 7.3937  \\
\addlinespace
\multirow{2}{*}{\textbf{D3.2}} 
    & \textbf{RMSE} & \textbf{8.9619} & 9.1044 & 9.6073 & 9.0432  \\
    & \textbf{MAE}  & \textbf{6.2427} & 6.3014 & 6.8097 & 6.2948  \\
\addlinespace
\midrule
\end{tabularx}
\end{table}

\begin{figure*}[!htbp]
    \centering
    \subfloat[RMSE in D1.1,D2.1 and D3.1]{%
        \label{fig:subfig_a}%
        \includegraphics[width=0.5\textwidth]{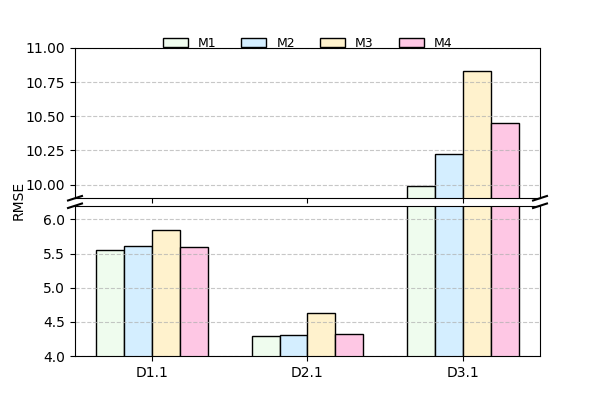}%
    }
    \hfill
    \subfloat[MAE in D1.1,D2.1 and D3.1]{%
        \label{fig:subfig_b}%
        \includegraphics[width=0.5\textwidth]{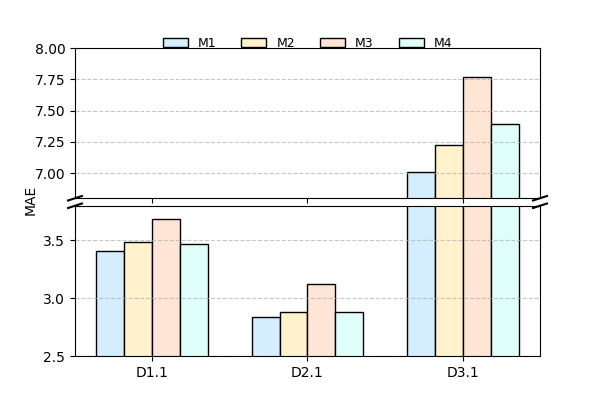}%
    }
     \caption{RMSE and MAE on the 10\% training set}
    \label{fig:rmse_mae_10_subfigs}
\end{figure*}

\begin{figure*}[!htbp]
    \centering
    \subfloat[RMSE in D1.2, D2.2 and D3.2]{%
        \label{fig:subfig_c}%
        \includegraphics[width=0.5\textwidth]{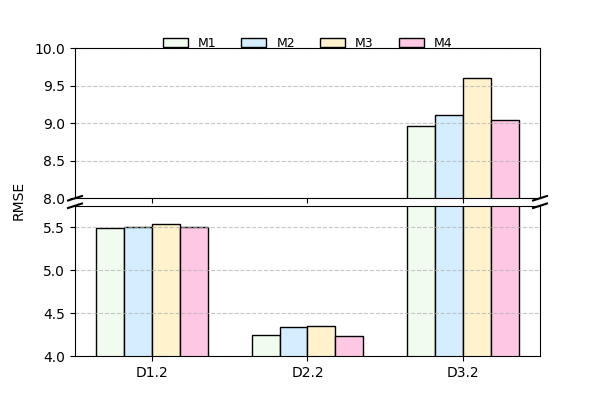}%
    }
    \hfill
    \subfloat[MAE in D1.2, D2.2 and D3.2]{%
        \label{fig:subfig_d}%
        \includegraphics[width=0.5\textwidth]{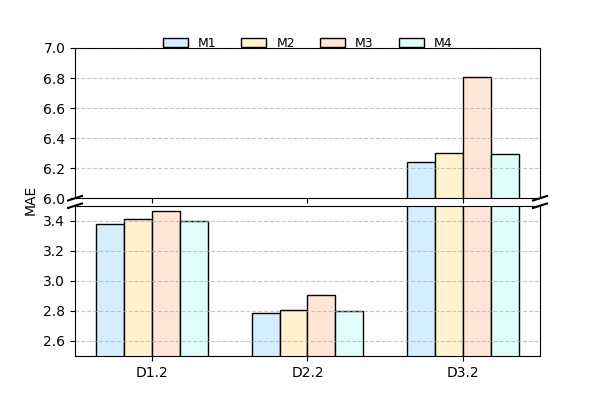}%
    }
    \label{fig:rmse_mae_20_subfigs}
    \caption{RMSE and MAE on the 20\% training set}
\end{figure*}

\textbf{(a) The RMSE and MAE achieved by TATSI are lower than those of the other three models on datasets D1.1, D2.1, and D3.1.} As shown in Table~\ref{tab:results} and Fig.~\ref{fig:rmse_mae_10_subfigs}, the minimum RMSE achieved by M1 on D1.1 is 5.5488, which is 1.19\% lower than M2’s (5.6154), 5.07\% lower than M3’s (5.8454), and 0.82\% lower than M4’s (5.5947), respectively. Similarly, the minimum MAE achieved by M1 on D1.1 is 3.4083, which is 2.14\% lower than M2’s (3.4829), 7.55\% lower than M3’s (3.6865), and 1.76\% lower than M4’s (3.4694). On D2.1, the minimum RMSE achieved by M1 is 4.2949, representing a 0.44\% reduction relative to M2’s (4.3139), a 7.27\% reduction relative to M3’s (4.6317), and a 0.53\% reduction relative to M4’s (4.3179). Additionally, the minimum MAE achieved by M1 on D2.1 is 2.8414, which is 1.34\% lower than M2’s (2.8799), 9.05\% lower than M3’s (3.1242), and 1.22\% lower than M4’s (2.8765). For D3.1, the minimum RMSE achieved by M1 is 9.9916, corresponding to a 2.24\% decrease compared to M2’s (10.2205), a 7.74\% decrease compared to M3’s (10.8299), and a 4.38\% decrease compared to M4’s (10.4492). Similarly, the minimum MAE achieved by M1 on D3.1 is 7.0067, which is 3.05\% lower than M2’s (7.2274), 9.78\% lower than M3’s (7.7656), and 5.23\% lower than M4’s (7.3937), respectively. These results demonstrate that TATSI achieves superior traffic speed prediction accuracy. A similar conclusion can be drawn from the other cases shown in Fig.~3.

\textbf{(b) As the proportion of training data increases, the TATSI model still demonstrates superior imputation accuracy.} Specifically, for dataset D1.2, the minimum RMSE achieved by M1 is 5.4963, which is 0.14\% lower than M2's 5.5039, 0.65\% lower than M3's 5.5322, and 0.08\% lower than M4's 5.5008. Similarly, the minimum MAE achieved by M1 on D1.2 is 3.3782, which is 1.03\% lower than M2's 3.4132, 2.43\% lower than M3's 3.4624, and 0.58\% lower than M4's 3.3980. For dataset D2.2, the minimum RMSE achieved by M1 is 4.2476, representing a 2.03\% reduction compared to M2's 4.3356 and a 2.44\% reduction compared to M3's 4.3540, although it is 0.23\% higher than M4's 4.2378. In terms of MAE, M1 attains a minimum value of 2.7841 on D2.2, which is 0.75\% lower than M2's 2.8052, 4.14\% lower than M3's 2.9043, and 0.45\% lower than M4's 2.7967. For dataset D3.2, the minimum RMSE achieved by M1 is 8.9619, which is 1.57\% lower than M2's 9.1044, 6.72\% lower than M3's 9.6073, and 0.90\% lower than M4's 9.0432. Moreover, the minimum MAE achieved by M1 on D3.2 is 6.2427, which is 0.93\% lower than M2's 6.3014, 8.33\% lower than M3's 6.8097, and 0.83\% lower than M4's 6.2948.

\section{Conclusion}
To accurately impute unobserved time-varying traffic speed data, this study proposes the Temporal-Aware Traffic Speed Imputation (TATSI) model. The model employs a combination of the \(SL_1\)-norm and \(L_2\)-norm in its loss function, along with robust regularization, which together confer both robustness and high accuracy in imputing missing data from an HDI tensor. Experimental results on three real-world time-varying traffic speed datasets demonstrate that TATSI significantly outperforms state-of-the-art models in prediction accuracy for missing data from an HDI tensor. Furthermore, developing an adaptive strategy for the hyperparameter \(\lambda\) presents an interesting challenge, which we plan to investigate in future work.

\bibliography{ref}
\end{document}